\documentclass[sigconf,nonacm,10pt]{acmart}

\AtBeginDocument{%
  }

\title{Translation from Wearable PPG to 12-Lead ECG}

\author{Hui Ji, Wei Gao, Pengfei Zhou}
\affiliation{%
\institution{University of Pittsburgh}
\city{Pittsburgh}
\state{PA}
\country{USA}
}

\acmConference[MobiCom '25]{The 31st Annual International Conference on
Mobile Computing and Networking}{Nov. 4-8, 2025}{Hong Kong, China}

\usepackage{algorithm}
\usepackage{algpseudocode}
\usepackage{makecell}
\usepackage{booktabs}
\usepackage{multirow}
\usepackage{xcolor}
\usepackage{balance}
\usepackage{natbib}
\usepackage{graphicx}
\usepackage{amsmath}  %

\usepackage{enumitem}  
\usepackage{float}

\usepackage{tikz}

\begin{document}

\title{Translation from Wearable PPG to 12-Lead ECG}


\begin{abstract}
The 12-lead electrocardiogram (ECG) is the gold standard for cardiovascular monitoring, 
offering superior diagnostic granularity and specificity compared to photoplethysmography (PPG). 
However, existing 12-lead ECG systems rely on cumbersome multi-electrode setups, 
limiting sustained monitoring in ambulatory settings, while current PPG-based methods 
fail to reconstruct multi-lead ECG due to the absence of inter-lead constraints and 
insufficient modeling of spatial-temporal dependencies across leads. 
To bridge this gap, we introduce P2Es, an innovative demographic-aware diffusion 
framework designed to generate clinically valid 12-lead ECG from PPG signals via 
three key innovations. Specifically, in the forward process, we introduce 
frequency-domain blurring followed by temporal noise interference to simulate 
real-world signal distortions. In the reverse process, we design a temporal 
multi-scale generation module followed by frequency deblurring. 
In particular, we leverage KNN-based clustering combined with contrastive learning 
to assign affinity matrices for the reverse process, enabling demographic-specific 
ECG translation. Extensive experimental results show that P2Es outperforms 
baseline models in 12-lead ECG reconstruction.
\end{abstract}
\maketitle



\section{Introduction}
\begin{figure}[t!]
  \centering
   \includegraphics[width=\linewidth]{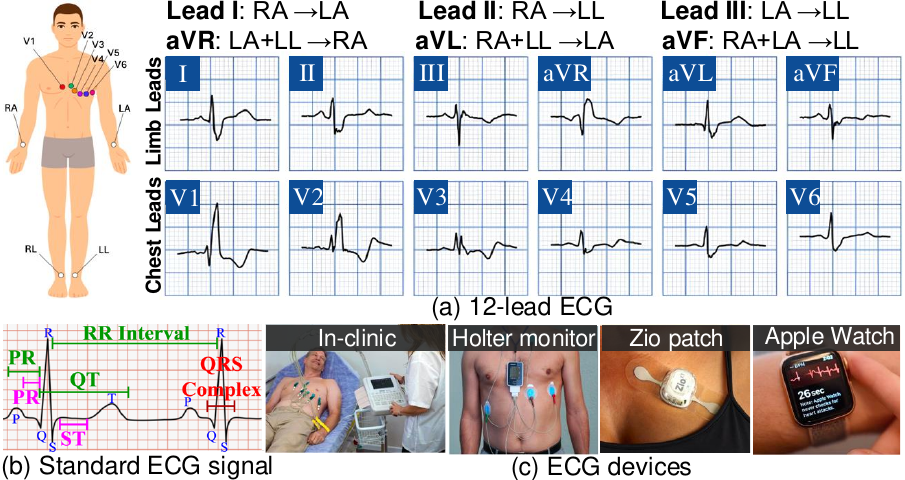}
   \vspace{-0.25in}
   \caption{(a) 12-Lead ECG. (b) Components in the standard ECG signal. (c) Different ECG devices.}
   \label{fig:f1}
   \vspace{-0.25in}
\end{figure}

In current clinical practices, the 12-lead electrocardiogram (ECG) is the gold standard for diagnosing cardiac abnormalities, providing comprehensive insights into cardiac electrical activity across the six limb leads (I, II, III, aVR, aVL, aVF) and six chest leads (V1–V6) \cite{tash2023cardiovascular}, as shown in Fig.~\ref{fig:f1}(a). It enables the detection of conditions such as myocardial ischemia, arrhythmia subtypes, and conduction disorders—issues that are often undetectable in other single-lead alternatives. 

However, widespread 12-lead ECG use, especially outside clinical settings, faces barriers as shown in Fig.~\ref{fig:f1}(c): Holter monitors are bulky, wired 12-lead devices restrict mobility, and hospital-grade setups are cost-prohibitive in low-resource settings \cite{lamichhane2022improved}. Although recent advances in wearable ECG monitors (e.g., Zio patch \cite{irhythm2025}) partially address portability, they still require users to manually place electrodes, limiting continuous monitoring \cite{guzik2016ecg} out of clinic. Wearable devices (e.g., Apple Watch and Fitbit wristbands) further simplify ECG acquisition but restrict outputs to single-lead formats (Lead I), sacrificing critical spatial patterns required for diagnosing complex cardiac conditions like myocardial ischemia or bundle branch blocks \cite{hong2019wearable}. 
They are also limited to intermittent ECG measurements,
where the user has to manually hold the finger on Apple Watch for 30 seconds for one ECG measurement \cite{ApplewatchECG}.
These limitations stem from a fundamental trade-off: reducing the count of electrodes erodes diagnostic specificity, but increasing the number of leads escalates the hardware complexity. Consequently, a critical question arises: \textit{To bridge this gap of ECG accessability, can diagnostically meaningful 12-lead ECG be generated from other physiological signals that are commonly available on wearables, such as PPG?}

Existing attempts to map physiological signals to multi-lead ECG face three unresolved challenges. \emph{\textbf{1)} Signal modality mismatch}: Prior works focus on lead-to-lead ECG synthesis (e.g., deriving V1–V6 from limb leads) \cite{seo2022multiple, yoon2024classification}, assuming the availability of at least one clinical-grade ECG input. These methods fail to address the challenge of cross-modality translation from PPG to ECG signals, which requires bridging two distinct signal representations. \emph{\textbf{2)} Anatomical inconsistency}: Existing learning frameworks (e.g., GANs and VAEs) generate ECG signals in their global latent spaces, but ignore demographic-specific variations in electrical axis orientation and chest lead correlations \cite{rodriguez2011recognition, villa2023unified}. This leads to implausible waveforms, such as inverted R-waves (see Fig.~\ref{fig:f1}(b)) in V2–V3 for anterior infarction patients \cite{moody1985derivation}. \emph{\textbf{3)} Spectral-temporal distortion}: While ECG features like ST-segment elevation (time-domain) and high-frequency notch waves (frequency-domain) demand co-optimization, current methods are limited to prioritizing only one domain \cite{ji2024advancing, christov2006comparative, stridh2004sequential}. For instance, wavelet-based models \cite{diery2009novel} preserve QRS complexes but blur P-waves, and temporal CNNs maintain waveform continuity at the cost of spectral fidelity \cite{basak2024novel}.

\begin{figure}[t!]
  \centering
   \includegraphics[width=0.8\linewidth]{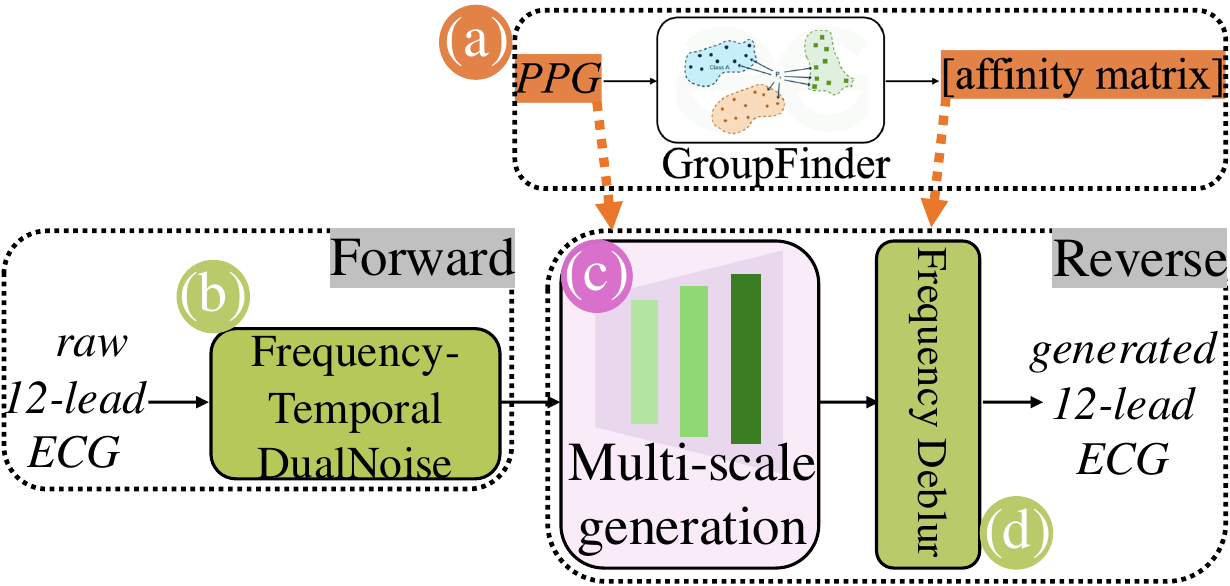}
   \vspace{-0.1in}
   \caption{The proposed P2Es framework.}
   \label{fig:sample}
   \vspace{-0.3in}
\end{figure}

To address these challenges, in this paper we present \emph{P2Es}, a new demographic-aware diffusion framework that generates 12-lead ECG from PPG through three novel AI mechanisms, as shown in Fig.~\ref{fig:sample}.

\noindent$\bullet$\hspace{0.5em}  \textbf{Demographic-aware Affinity Modeling:} Unlike prior clustering-based methods that rely solely on static demographics \cite{latha2021fully, castineira2020adding}, we design a dynamic GroupFinder module (Fig. \ref{fig:sample}(a)) that combines KNN-based clustering with contrastive learning. This module jointly optimizes two objectives: (i) Demographic similarity, encompassing attributes such as age, gender, and medical records, to group patients with shared anatomical traits (e.g., ventricular hypertrophy prevalence in elderly males); and (ii) Waveform affinity, leveraging PPG-ECG correlation to cluster subjects with hemodynamic similarities (e.g., pulse transit time variability in hypertensive patients). The resulting subgroup-specific affinity matrices encode inter-lead voltage constraints and dynamically adjust during the diffusion process, ensuring physiological consistency among different ECG leads. 

\noindent$\bullet$\hspace{0.5em}  \textbf{Frequency-Temporal Diffusion:} Breaking from conventional noise scheduling, we design a dual-phase forward process, Frequency-Temporal DualNoise (Fig. \ref{fig:sample}(b)): Frequency-domain blurring via Gaussian filter banks simulates spectral degradation, followed by time-domain noise injection. The dual-phase reverse process cascades temporal multi-scale generation (Fig. \ref{fig:sample}(c)) and frequency deblur (Fig. \ref{fig:sample}(d)) modules. This dual-domain diffusion outperforms single-domain methods by resolving conflicts (e.g., preserving ST-segment slope while suppressing high-frequency ringing).

\noindent$\bullet$\hspace{0.5em}  \textbf{Temporal Multi-Scale Generation:} To capture amplitude variations across ECG leads, we implement a multi-scale reverse process (Fig.\ref{fig:sample}(c)) that progressively refines temporal features at different noise levels, leveraging dilated convolutions and attention gates. At coarse noise levels, large-kernel dilated convolutions capture overall trends, such as baseline wander. At medium noise levels, attention gates guided by the affinity matrix prioritize ST segments in chest leads and P-waves in limb leads, whereas at fine noise levels, high-frequency noise enhances the local resolution of the QRS notch and J point, accurately preserving microvolt-level pathological features. This hierarchical approach ensures that low-amplitude features are preserved without being overshadowed by dominant R-peaks.

To enable diffusion-based ECG generation on mobile devices, our design optimizes the compute efficiency through adaptive sampling, parameter sharing, and input length selection. DDIM-based \cite{song2020denoising} sampling accelerates inference by adjusting step sizes for noise removal and detail refinement. A parameter-sharing design reduces redundancy across diffusion steps and the 12-lead generation network. Input length is optimized to balance performance, computational cost, and latency for real-time deployment.

P2Es has been evaluated on three datasets: MIMIC-II (training the GroupFinder module), MIMIC-III (performance benchmarking), and MIMIC-IV (performance benchmarking and clinical validation). P2Es achieves state-of-the-art (SOTA) performance compared to PPG-to-single-lead baselines, where we reduce reconstruction errors by \textbf{16.3\%} in MSE (0.0902 vs. 0.1078) and \textbf{26.4\%} in DTW (0.0583 vs. 0.0792). P2Es also achieves the best performance when compared with the two-stage generation approaches (i.e., PPG to single-lead ECG and single-lead ECG to 12-lead ECG) based on existing models.
For cardiovascular disease diagnosis on MIMIC-IV, 12-lead ECG generated by P2Es achieve \textbf{87.5\% specificity} in detecting myocardial infarction and \textbf{83.5\% precision} in arrhythmia classification. 

Our main contributions are summarized as follows:
\begin{itemize}[leftmargin=*]
\setlength{\itemsep}{0pt}  
    \item P2Es is the first framework to generate 12-lead ECG directly from wearable PPG signals, bridging the modality gap through diffusion-based cross-domain alignment.
    \item We introduce three innovations to realize P2Es. First, a hybrid KNN-contrastive learning module is designed to derive demographic-aware affinity matrices to guide the reverse process. Second, the frequency-temporal diffusion with dual-domain noise injection preserves spectral coherence while resisting temporal distortions. Third, the temporal multi-scale hierarchical reverse process employs cascaded dilated convolutions with expanding receptive fields to achieve global rhythm stabilization, local morphological enhancement, and high-frequency detail recovery.
    \item We propose the mobile-first lightweight design that enables real-time 12-lead ECG generation on mobile devices. We prototyped P2Es on smartphones and evaluated its performance on three datasets and compared with various baselines and existing methods. P2Es achives the best performance in all settings and has been validated in cardiovascular disease diagnosis tasks.
\end{itemize}

\section{Background and Challenges}
\subsection{Multi-Lead ECG and Wearable PPG}
\begin{figure}[t!]
  \centering
   \includegraphics[width=1\linewidth]{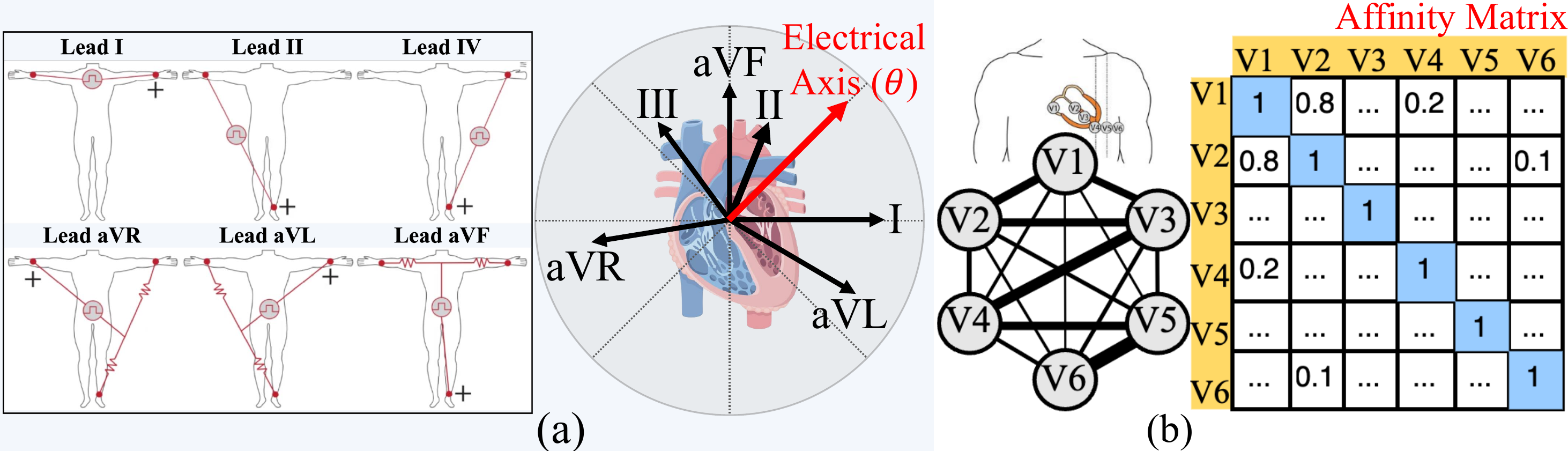}
   \vspace{-0.28in}
   \caption{(a). Electrical axis in limb leads. (b). Principle of affinity matrix in chest leads.}
   \label{fig:EAAM}
   \vspace{-0.28in}
\end{figure}

The 12-lead ECG captures cardiac electrical activity across multiple anatomical planes, and has been the gold standard for cardiovascular diagnosis for decades. It comprises six limb leads (I, II, III, aVR, aVL, aVF) and six chest leads (V1–V6). Limb leads derive from three electrodes (RA, LA, LL), forming Einthoven’s triangle (Fig.~\ref{fig:EAAM} (a)), with Lead I (RA $\rightarrow$ LA), Lead II (RA$\rightarrow$LL), and Lead III (LA$\rightarrow$LL). Augmented leads (aVR, aVL, aVF) enhance sensitivity to specific axes via Wilson’s central terminal. Chest leads (V1–V6) are placed sequentially across the chest, with spatial correlations quantified by an affinity matrix (Fig.~\ref{fig:EAAM} (b)). High coefficients (e.g., V4–V5: 0.8) indicate overlapping myocardial regions. This standardized setup enables precise diagnosis: limb leads assess frontal plane electrical axis ($\theta$) for axis deviation, while chest leads localize transverse plane pathologies. 

Photoplethysmography (PPG), in contrast, measures peripheral blood volume fluctuations via optical sensors, offering non-invasive and continuous monitoring. Its widespread adoption in wearable devices stems from advantages such as low cost, ease of use, and the ability to capture cardiovascular responses during everyday activities \cite{almarshad2022diagnostic}. However, PPG signals are inherently limited by their dependence on hemodynamic coupling (e.g., pulse transit time) rather than direct electrophysiological data. This restricts diagnostic utility to basic rhythm analysis (e.g., atrial fibrillation detection) and heart rate estimation. Crucially, PPG cannot localize pathologies (e.g., distinguishing anterior vs. inferior infarction) due to the absence of multi-lead spatial correlations \cite{shabaan2020survey}.

The urgent need to translate PPG into diagnostic-grade 12-lead ECG is driven by two global imperatives: First, over 80\% of cardiovascular deaths occur in low-resource regions lacking ECG infrastructure \cite{khurshid2023electrophysiology}, where PPG-enabled wearables could democratize access to cardiac diagnostics; second, traditional ECG provides only episodic snapshots, whereas PPG supports long-term surveillance essential for detecting transient events like silent ischemia.

\begin{figure}[t!]
  \centering
   \includegraphics[width=1\linewidth]{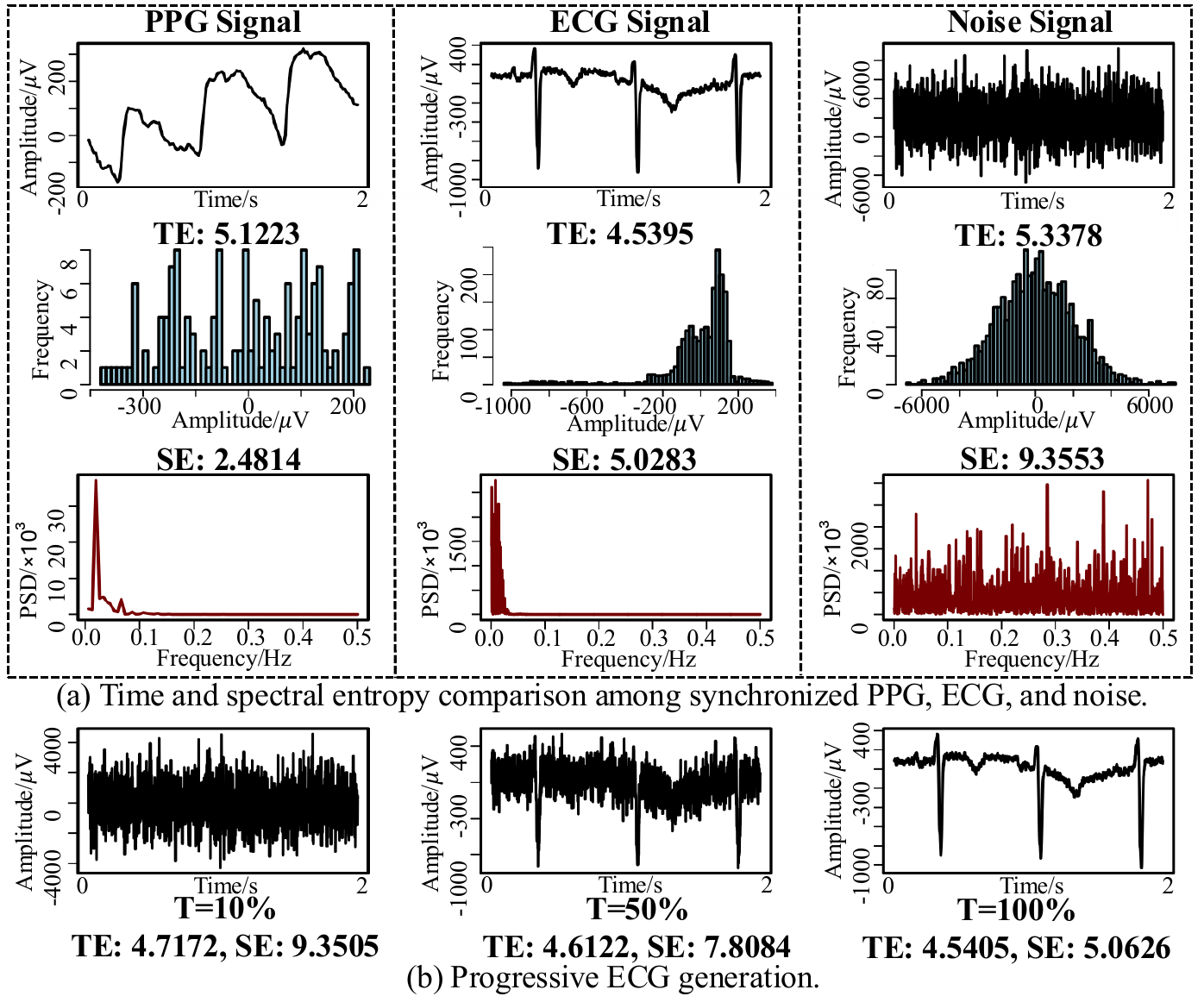}
   \vspace{-0.25in}
   \caption{Information entropy analysis (TE: time-domain entropy; SE: spectral entropy).}
   \label{fig:entropy}
   \vspace{-0.25in}
\end{figure}

\vspace{-0.1in}
\subsection{Information-Theoretic Feasibility}
To illustrate the theoretical feasibility of transforming PPG into ECG, we use the V1 lead ECG as an example and analyze the intrinsic information entropy of the signal. Fig.\ref{fig:entropy}(a) compares the time-domain amplitude histogram and frequency-domain power spectrum of a synchronized PPG segment, a V1 lead ECG segment, and a noise signal, which is obtained by progressively adding noise to the V1 lead ECG in both the time and frequency domains until it becomes white noise.

We estimate the amplitude-domain entropy using the standard Shannon entropy formula~\cite{cover1999elements}.
Given a continuous signal $X$, its time-domain entropy (TE) is computed by discretizing its amplitude 
into $K$ bins: $H_{\text{time}}(X) = - \sum_{i=1}^{K} p(x_i) \log p(x_i)$, 
where $p(x_i)$ is the probability that a sample falls into bin $i$.
The spectral entropy (SE) follows~\cite{inouye1991quantification} and is computed 
using the normalized power spectral density (PSD) as 
$H_{\text{spectral}}(X) = -{\textstyle \sum_{j=1}^{M}}\, p(f_j) \log p(f_j)$.
As shown in Fig.\ref{fig:entropy}(a), the ECG signal has the lowest time-domain entropy (TE=4.54), reflecting its highly repetitive P-QRS-T morphology. In contrast, PPG has higher time-domain entropy (TE=5.12), suggesting that its amplitude variation covers a wider dynamic range than ECG.
Meanwhile, the spectral entropy of PPG (SE=2.48) is lower than that of ECG (SE=5.03), indicating that PPG primarily preserves dominant low-frequency cardiac rhythms, but lacks the richer high-frequency harmonics found in ECG (e.g., sharp QRS onsets). 
This mutual constraint highlights that PPG provides valuable priors for ECG recovery in both the time and frequency domains, but the information it contains in each domain is inherently different. Therefore, effective PPG-to-ECG generation must combine and align both domains, as relying on either one alone is insufficient to fully recover the structured ECG signal.

According to information theory~\cite{cover1999elements}, conditioning on additional observations reduces uncertainty, so $H(Y|X) \leq H(Y)$. Here, $Y$ represents the target ECG signal to be recovered from pure noise in the unconditional setting, and $X$ is the observed PPG. In the time domain, PPG’s moderately lower entropy than noise but higher than ECG provides useful structure to guide ECG toward a more ordered waveform. In the frequency domain, PPG’s lower spectral entropy means it retains core low-frequency rhythms, acting as an anchor for ECG’s dominant frequencies while high-frequency details must be reconstructed. This shows that PPG as a conditional input is not only feasible but necessary: without it, generating ECG from pure noise would leave an overly large and uncertain search space, whereas the PPG provides complementary time- and frequency-domain anchors to reduce posterior entropy.


\subsection{How Diffusion Reduces Uncertainty}
Diffusion models~\cite{croitoru2023diffusion, song2020denoising} provide a natural framework for cross-modal time-series generation. The forward process adds noise to structured ECG signals, increasing entropy until the data approximates pure noise. The reverse process then denoises this high-entropy signal step by step, recovering structured low-entropy ECG waveforms.
Fig.~\ref{fig:entropy}(b) visualizes entropy reduction during reverse diffusion at representative steps: 10\%, 50\%, and 100\%. At T=10\%, the output remains noise-like (SE=9.35). By T=50\%, cardiac harmonics emerge (SE=7.81), and by T=100\%, the model generates a physiologically plausible waveform (SE=5.06), close to the ground truth. This shows that diffusion model does not hallucinate but systematically reduces uncertainty via conditional PPG guidance and learned physiological priors.
Compared to GANs~\cite{durgadevi2021generative} and VAEs~\cite{bond2021deep}, which often struggle with mode collapse or blurry outputs, diffusion models preserve temporal structure and exhibit strong noise robustness—critical for real-world PPG signals~\cite{lin2024diffusion}. Yet, prior work is limited to unimodal or single-lead ECG generation~\cite{ji2024advancing}, leaving cross-modal, multi-lead synthesis underexplored. 

\subsection{Challenges}
Despite the information-theoretic feasibility and empirical evidence of stepwise entropy reduction, practical cross-modal 12-lead ECG generation still faces key challenges. Translating PPG signals to diagnostic-grade multi-lead ECG is physiologically complex because PPG measures optical blood volume, whereas ECG reflects electrical cardiac activity shaped by nonlinear hemodynamics~\cite{miller1988relationship, garcia200112, pipberger1961correlation}. This mismatch makes direct mapping difficult, and existing methods often ignore inter-lead dependencies and physiological consistency (e.g., Einthoven’s triangle~\cite{zhao2022einthoven}), leading to anatomically implausible outputs. While PPG strongly constrains overall cardiac rhythm, it does not capture detailed spatial voltage relationships among all ECG leads, which are critical for recovering lead-specific features. Faithful joint time-frequency representation is also challenging: ECG requires both precise timing (e.g., ST-segment alignment~\cite{pilgrim2016risk}) and frequency coherence up to high-frequency QRS harmonics~\cite{sornmo2005bioelectrical, martinek2021advanced}, yet many models still favor one domain, producing distorted waveforms~\cite{balaji2018power}. Finally, multi-lead ECG synthesis must handle substantial amplitude differences—limb leads typically range from 0.1–2~mV, chest leads from 0.5–5~mV~\cite{proniewska2023deltawaveecg, khunti2014accurate}—while remaining lightweight enough for wearable use without sacrificing signal quality.

\section{Proposed Method}
\subsection{Overview}
\begin{figure*}[t]
 \centering
  \includegraphics[width=0.85\linewidth]{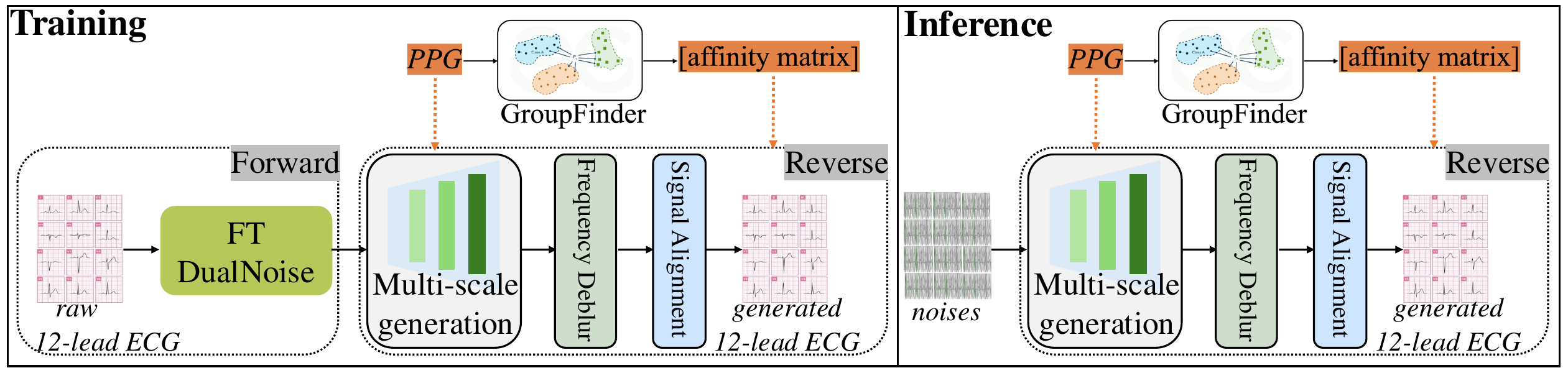}
  \caption{
  Overview of the P2Es framework: training consists of a forward process (FT DualNoise) and a reverse process (guided by PPG and the affinity matrix) for ECG reconstruction via multi-scale generation, frequency deblurring, and alignment. Inference contains only the reverse process.}
  \label{fig:framework}
\end{figure*}

To solve challenges above, we propose P2Es, a demographic-aware time-frequency diffusion model as shown in Fig.~\ref{fig:framework}. During the training phase, the forward process begins with the original 12-lead ECG signals, introducing noise in both frequency and time domains through the Frequency-Temporal (FT) DualNoise module. In the reverse process, the noise undergoes multi-scale generation, frequency-domain deblurring, and signal alignment. The final 12-lead ECG signals are generated by conditioning on the PPG signals and their corresponding affinity matrices, which are obtained via the GroupFinder module. In the inference phase, only the PPG signal and associated affinity matrix are required as inputs.


\subsection{GroupFinder and Affinity Matrix}
Prior to training, we perform KNN-based clustering on the MIMIC II dataset to segment subjects into different groups. Each group is assigned an affinity matrix $A_g$, which conditions the reverse diffusion process, ensuring that ECG reconstruction aligns with demographic-aware characteristics.

\subsubsection{Affinity Matrix Construction}
The affinity matrix quantifies spatial correlations between chest leads (V1-V6), essential for preserving anatomical plausibility in generated multi-lead ECG. For each subject $i$, we compute the Pearson correlation coefficient \( A_i(m,n) \):
\begin{equation}
  \mathbf{A}_i(m,n) = \frac{\mathbb{E}[(V_m^{(i)} - \mu_{V_m})(V_n^{(i)} - \mu_{V_n})]}{\sigma_{V_m}\sigma_{V_n}},\, 1 \leq m,n \leq 6.
\end{equation}
The normalized signal of lead \( V_m \) for the \( i \)th subject is represented as \( V_m^{(i)} \). Additionally, \( \mu_{V_m} \) and \( \sigma_{V_m} \) denote the mean and standard deviation of lead \( V_m \), respectively. This formulation extends Kornreich's anatomical correlation analysis \cite{kornreich1991identification} to dynamic demographic characteristics. When \( m = n \), the numerator is the variance of \( V_m \), and the denominator is \( \sigma^2_{V_m} \), thus \( A_i(m,m) = 1 \) (self-correlation is 1). This is expected because signals from the same connected lead are perfectly correlated. The 6×6 matrix specifically captures (i) anterior-posterior voltage gradients (V2-V4 correlations), (ii) right-left ventricular activation patterns (V1-V5 anti-correlations) and (iii) ST-segment spatial consistency (simultaneous elevation/depression).

\begin{figure}[t!]
  \centering
   \includegraphics[width=1.0\linewidth]{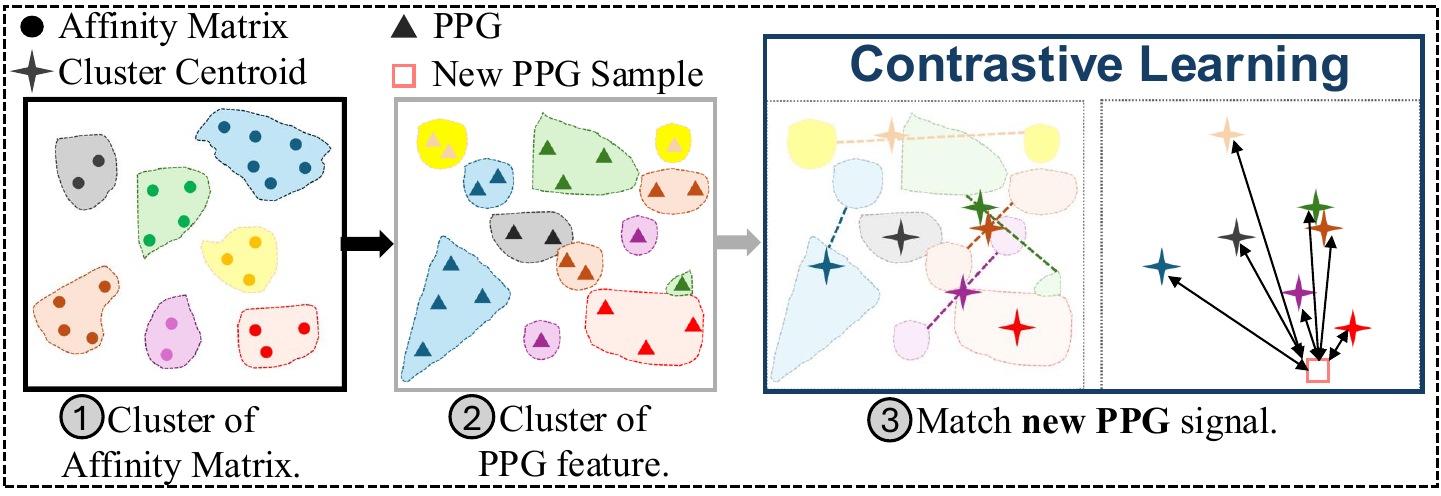}
   \vspace{-0.2in}
   \caption{Details of the GroupFinder Module.}
   \vspace{-0.2in}
\end{figure}

\subsubsection{KNN-Based Clustering}
To dynamically segment subjects into subgroups that jointly reflect demographic similarity and affinity matrix, we propose a multi-modality KNN clustering framework. The process is structured as follows:
  Flatten upper triangle elements of $\mathbf{A}_i$ into $\mathbf{a}_i \in \mathbb{R}^{21}$, preserving spatial symmetry:
  \begin{equation}
    \mathbf{a}_i = [\mathbf{A}_i(1,1), \mathbf{A}_i(1,2), ..., \mathbf{A}_i(5,6), \mathbf{A}_i(6,6)].
  \end{equation}
Demographic similarity extracts demographic attributes \( \mathbf{d}_i \in \mathbb{R}^{p} \), including age, gender, medical flags (e.g., diabetes, hypertension), and ventricular hypertrophy status.
The final joint feature vector is:
\begin{equation}
    \mathbf{a}_i^{\text{joint}} = \left[ \mathbf{a}_i \,\, \Vert \,\, \mathbf{d}_i \right] \in \mathbb{R}^{21 + p},
\end{equation}
where \( \Vert \) denotes concatenation, and \( p \) is the dimension of demographic features.

\textbf{Cluster Optimization}: Next, we select the most appropriate number of group through Multi-criteria validation. For each sample \( i \), it is defined as:
\begin{equation}
    s(i) = \frac{b(i) - a(i)}{\max\{a(i), b(i)\}},
\end{equation}
where \( a(i) \) is the average intra-cluster distance (cohesion), and \( b(i) \) is the smallest average inter-cluster distance (separation). The global score \( S(K) \) is the mean of all \( s(i) \), that is, $S(K) = \frac{1}{N} \sum_{i=1}^{N} s(i)$. Here, \( S(7) = 0.62 \) (vs. 0.51 for \( S(K)  = 5)\) indicates well-defined clusters with strong intra-group similarity and inter-group distinction, justifying \( K = 7 \) as the optimal choice.



\textbf{Centroid Calculation}: For each cluster \( C_k \), we compute centroids as follows:
  \begin{equation}
    \bar{\mathbf{A}}_k = \frac{1}{|\mathcal{C}_k|} \sum_{i \in \mathcal{C}_k} \mathbf{a}_i^{\text{joint}}.
    \label{eq:centroid}
  \end{equation}


  
  


\subsection{Contrastive Assignment for New PPG Inputs}
\label{subsec:assignment}
To dynamically associate new PPG signals with predefined ECG clusters derived from joint affinity matrices, we propose a contrastive learning framework, structured as follows.

\subsubsection{Dual-Branch Encoder Architecture}

The PPG encoder $\mathcal{E}_{\text{PPG}}$ maps raw PPG signals $x_i$ and features $f_i = [A_s, T_{PA}, S_d]$ to an embedding space: $z_i^{\text{PPG}} = \mathcal{E}_{\text{PPG}}(x_i) \in \mathbb{R}^{d_e}.$ Here, $A_s$, $T_{PA}$, $S_d$ \cite{park2022photoplethysmogram} represent the peak amplitude of the systolic phase in the PPG signal, the time delay from the R wave to the PPG systolic peak, and the PPG diastolic slope, respectively. The affinity matrix encoder $\mathcal{E}_{\text{A}}$ encodes cluster centroids $\bar{A}_k$ into the same space: $c_k = \mathcal{E}_{\text{A}}(\bar{A}_k) \in \mathbb{R}^{d_e}.$

\subsubsection{Contrastive Loss Function}
The model minimizes the following loss to align PPG embeddings with their corresponding ECG clusters:
\begin{equation}
    \mathcal{L}_{\text{cont}} = -\log \frac{\exp(z_i^{\text{PPG}} \cdot c_k / \tau)}{\sum_{m=1}^7 \exp(z_i^{\text{PPG}} \cdot c_m / \tau)},
\end{equation}
where $\tau$ is a hyperparameter. A lower \(\tau\) makes the model focus more on the most similar matches, while a higher \(\tau\) smooths the distribution.
\label{subsec:assignment}
For a new PPG signal $x_{\text{new}}$, the cluster assignment proceeds as Algorithm \ref{algo1}.



\begin{algorithm}[t]
\caption{Real-Time Contrastive Group Assignment}
\label{algo1}
\begin{algorithmic}[1]
\Require New PPG signal $x_{\text{new}}$, pre-trained encoders $\mathcal{E}_{\text{PPG}}$ and  $\mathcal{E}_{\text{A}}$, cluster centroids $\{c_k\}_{k=1}^{7}$.
\Ensure Assigned affinity matrix $\bar{A}_{k^*}$.

\State Extract features from $x_{\text{new}}$: $f_{\text{new}} = [A_s, T_{PA}, S_d]$.
\State Embed features using encoder $\mathcal{E}_{\text{PPG}}$: $z_{\text{new}} = \mathcal{E}_{\text{PPG}}(f_{\text{new}})$.
 
\For{$k = 1$ to $7$} 
    \Statex Compute cosine similarity:
    \[
    s_k = \frac{z_{\text{new}} \cdot c_k}{\|z_{\text{new}}\|\|c_k\|}
    \]
\EndFor

\State Identify optimal cluster index: 
\[
k^* = \arg\max_k s_k
\]
\State Retrieve affinity matrix $\bar{A}_{k^*}$ for diffusion conditioning.
\end{algorithmic}
\end{algorithm}
\vspace{-8pt}

\subsection{Forward Process: Frequency Blurring and Temporal Noise}
During forward process, FT DualNoise model integrates frequency-domain blurring and QRS-adaptive temporal noise injection to balance spectral degradation and morphological preservation. This two-stage process ensures that critical cardiac features (e.g., QRS complexes) remain identifiable. 

\subsubsection*{Stage 1: Frequency Blurring Module}
Given a 12-lead ECG signal $\mathbf{x}_0 \in \mathbb{R}^{N \times 12}$ ($N$: time steps), we first transform it to the frequency domain via the Discrete Fourier Transform (DFT):
\begin{equation}
    \mathbf{X}_0 = \mathcal{F}[\mathbf{x}_0] = U \mathbf{x}_0 \in \mathbb{C}^{N \times 12},
\end{equation}
where $U \in \mathbb{C}^{N \times N}$ is the unitary DFT matrix. To simulate spectral distortion, we apply a frequency blur operator $\mathcal{B}$:
\begin{equation}
    \mathbf{X}_{\text{blur}} = \mathcal{B}(\mathbf{X}_0) = \mathbf{X}_0 \odot M + \epsilon_f,
\end{equation}
where $M \in \mathbb{R}^{N \times 12}$ means Gaussian low-pass mask with bandwidth $\omega_c = \frac{\pi t}{T}\cdot f_{\max}$, where $f_{\max}$ is the maximum frequency of the signal, $t$ is the diffusion timestep and $T$ the total steps. $\epsilon_f \sim \mathcal{CN}(0, \sigma_f^2 I)$ is complex Gaussian noise added to high-frequency components.
The blurred signal is then mapped back to the time domain:
\begin{equation}
    \mathbf{x}_{\text{blur}} = \mathcal{F}^{-1}[\mathbf{X}_{\text{blur}}] = U^* \mathbf{X}_{\text{blur}}.
\end{equation}

\subsubsection*{Stage 2: QRS-Adaptive Temporal Noise}
In order to further capture the most important QRS area in the ECG signal, the blurred signal $\mathbf{x}_{\text{blur}}$ then undergoes temporal noise injection, modulated by QRS complexes detected via an adaptive search.
\textbf{QRS Mask Generation.} Using a modified Pan-Tompkins algorithm \cite{pan1985real}, we generate a binary mask $\mathbf{m} \in \{0,1\}^{N \times 12}$:
\begin{equation}
    m_{\tau,j} = \begin{cases} 1, & \text{if } \tau \in \text{QRS region of lead } j; \\
    0, & \text{otherwise.} \end{cases}
\end{equation}

\noindent \textbf{Temporal Noise Injection.} At diffusion timestep $t$, noise is selectively added to non-QRS regions:
\begin{equation}
    \mathbf{x}_t = \sqrt{\bar{\alpha}_t} \mathbf{x}_{\text{blur}} + \sqrt{1 - \bar{\alpha}_t} (1 - \mathbf{m}) \odot \epsilon_t,
\end{equation}
where $\bar{\alpha}_t = \prod_{s=1}^{t} (1 - \beta_s)$, and \( \delta_t \) represents a small positive constant acquired from a fixed variance schedule \( \delta_1, \delta_2, ..., \delta_t \). Here $\epsilon_t \sim \mathcal{N}(0, I)$ means standard Gaussian noise and $\odot$ is Hadamard product.





\subsection{Reverse Process: Multi-Scale Temporal Recovery and Frequency Debluring}
As illustrated in Fig. \ref{fig:framework}, the reverse process, guided by PPG signals and affinity matrices, reconstructs 12-lead ECG through multi-scale denoising, followed by frequency-domain deblurring and signal alignment. This hierarchical strategy ensures both global waveform fidelity and local morphological precision, leveraging domain-specific priors. 

\textbf{Multi-Scale Temporal Recovery.} Given T is the total step of the diffusion model, the noisy signal $\mathbf{x}_T$ from the forward process, the multi-scale temporal reverse iteratively refines the signal across three stages:

\subsubsection*{1. Early-Stage Coarse Denoising ($t \in [0, T/3]$)}
Dilated convolutions with kernel sizes $(9 \times 1)$ are used to capture long-range dependencies.
\begin{equation}
    \mathbf{x}_{t-1} = \frac{1}{\sqrt{\alpha_t}} \left( \mathbf{x}_t - \frac{1 - \alpha_t}{\sqrt{1 - \bar{\alpha}_t}} \epsilon_0(\mathbf{x}_t, t, \mathbf{c}) \right) + \sigma_t \mathbf{z},
\end{equation}
where $\mathbf{c} = \{\text{PPG}, \bar{\mathbf{A}}_{k}\}$ concatenates PPG features and the assigned affinity matrix, and $\mathbf{z} \sim \mathcal{N}(0, I)$. 

\subsubsection*{2. Mid-Stage Structural Refinement ($t \in [T/3, 2T/3]$)}
Hybrid layers combine convolutions $(7 \times 1)$ with transformer attention:
\begin{equation}
    \text{Attention}(\mathbf{Q}, \mathbf{K}, \mathbf{V}) = \text{softmax} \left( \frac{\mathbf{Q} \mathbf{K}^T}{\sqrt{d_k}} \right)\mathbf{V},
\end{equation}
where $\mathbf{Q} = \mathbf{x}_t W_Q$, $\mathbf{K} = \mathbf{x}_t W_K$, and $\mathbf{V} = \mathbf{x}_t W_V$ are learnable weight matrices that project the input into different semantic spaces.

\subsubsection*{3. Late-Stage Detail Recovery ($t \in [2T/3, T]$)}
 Stacked convolutions $(5 \times 1)$ and residual connections reconstruct QRS, P-waves, and T-waves with sub-millisecond precision.


\textbf{Frequency-Domain Deblurring.} After temporal denoising, the signal undergoes spectral restoration to counteract forward-process blurring. Here, assuming that the multi-scale restored signal is $\mathbf{x}^{\text{pred}}$:
\begin{equation}
    \mathbf{X}^{\text{pred}} = \mathcal{F}[x^{\text{pred}}] \in \mathbb{C}^{N \times 12},
\end{equation}
where \( N \) is the signal length, and each column corresponds to a lead's spectral representation. To amplify diagnostically critical high-frequency components (\(>15\) Hz), a frequency-selective mask \( M_{\text{high}} \) is applied:  
\begin{equation}
    \mathbf{X}^{\text{deblur}} = \mathbf{X}^{\text{pred}} \odot (1 + \gamma M_{\text{high}}),
\end{equation}
where $M_{\text{high}} \in \mathbb{R}^{N \times 12}$ is a bandpass mask for high-frequency components. $\gamma$ is a learnable parameter that controls the strength of the high-frequency enhancement. Next is the inverse discrete Fourier transform to recover the final signal:
\begin{equation}
    \mathbf{x}^{\text{final}} = \mathcal{F}^{-1}[\mathbf{X}^{\text{deblur}}].
\end{equation}

\textbf{Signal Alignment.}
In order to further improve the quality of reconstructed 12-lead ECG signals, we adopted a signal alignment module to align the signal amplitude and frequency \cite{ji2024advancing}. 
The loss function is calculated as:
\begin{equation}
    \mathcal{L}_a = ( \frac{1}{N_{\min}} \sum_{i=1}^{N_{\min}} \left| p_{g_i} - p_{t_i} \right| )
+ \left| E_g - E_t \right| + \left| N_g - N_t \right|,
\end{equation}
where \( N_{\min} \) is the minimum number of common R-peaks between generated and original multi-lead ECG, \( p_{g_i} \) and \( p_{t_i} \) represent the positions of the \( i \)-th R-peak in the generated and ground-truth signals, respectively. The amplitude ranges \( E_g \) and \( E_t \), calculated as \( \max(x) - \min(x) \), quantify peak-to-peak voltage differences in the generated and original multi-lead ECG. Heart rate consistency is measured by \( N_g \) and \( N_t \), derived from R-peak counts per unit time for the synthesized and reference signals.

\textbf{Unified Loss Function.} In our proposed P2Es, the overall
training objective is summarized as follows:
\begin{equation}
    \mathcal{L}_{\text{total}} = \lambda_1 \mathcal{L}_{\text{L}} + \lambda_2 \mathcal{L}_{\text{F}} + \lambda_3 \mathcal{L}_{\textbf{S}_{\theta}} + \lambda_4 \mathcal{L}_a.
\end{equation}

Among this,
\begin{equation}
    \mathcal{L}_{L} = \sum_{i \neq j} \left\| \Delta ECG_{i,j}^{Pred} - \Delta ECG_{i,j}^{GT} \right\|_2,
\end{equation}
\begin{equation}
    \mathcal{L}_{F} = \left\| \mathcal{F}(ECG^{GT}) - \mathcal{F}(ECG^{Pred}) \right\|_2,
\end{equation}
\begin{equation}
    \mathcal{L}_{\textbf{S}_{\theta}} = \left\| \mathbf{S}_{\theta}^{Pred} - \mathbf{S}_{\theta}^{GT} \right\|_2,
\end{equation}
with $\lambda_1 = 1.0$, $\lambda_2 = 0.5$, $\lambda_3 = 0.2$, $\lambda_4 = 0.3$.

\subsection{Lightweight Model Design}
The integration of PPG to 12-lead ECG generation into mobile device-based health monitoring systems demands stringent constraints on computational efficiency, memory footprint, and power consumption. Below we detail how our lightweight design addresses these challenges while maintaining clinical-grade accuracy.
\begin{itemize}[leftmargin=*]
\item \textbf{Efficient Sampling with DDIM \cite{song2020denoising}.} The 50-step DDIM reduces inference time by 81\% compared to standard 1000-step DDPM while preserving diagnostic features: Early steps (1-15) use large strides ($\Delta t = 5$) for rapid noise removal; late steps (16-50) refine details with ($\Delta t = 1$). 
\item \textbf{Parameter-sharing.} The P2Es adopts parameter-sharing, where all diffusion steps share the PPG encoder and affinity matrix projection MLP parameters. The 12-lead generation network shares the same set of convolution kernels, and only the input splicing lead number is embedded.
\item \textbf{Optimized Signal Length.} Additionally, to optimize our model for mobile-device inference, we explored various input signal lengths, evaluating their performance, computational cost, and latency trade-offs. We ultimately selected a signal length of 2 seconds as the optimal balance. 
\end{itemize}

Ultimately, the number of parameters is reduced by 28\%, and the inference speed is increased by 1.6 times.

\section{Experiments}

\subsection{Datasets and Experiment Details}
{\bfseries Datasets.}
We use the MIMIC series (II–IV) datasets~\cite{mehrgardt2022pulse, johnson2016mimic, johnson2023mimic}, compiled by MIT and collected at Beth Israel Deaconess Medical Center (BIDMC), Boston. These datasets span 2001–2019 and provide anonymized, high-resolution physiological waveforms primarily from ICU and emergency departments. MIMIC II includes signals such as non-invasive blood pressure (NIBP), PPG, and up to 12-lead ECG. MIMIC III extends this with additional waveforms including arterial blood pressure (ABP) and respiratory data. MIMIC IV further incorporates hospital-wide clinical records and continues to provide comprehensive ECG, PPG, ABP, and respiratory signals. 
PPG is recorded via bedside pulse oximeter probes (mainly Philips IntelliVue ~\cite{philips2023}) typically attached to the fingertip. Similar PPG sensors are now common in consumer wearables, such as smart rings ~\cite{oura2024, ultrahuman2024}, enabling continuous non-invasive monitoring beyond the ICU. Together, the MIMIC II–IV datasets offer the largest open-access, clinically annotated 12-lead ECG repository with paired PPG and diagnostic waveforms, serving as a benchmark for AI-driven cardiac research~\cite{saeed2011multiparameter, strodthoff2024prospects}.

{\bfseries Data pre-processing.} 
Patient identifiers across all three datasets were de-identified via randomized cryptographic mapping, generating consistent integer IDs for patients, hospitalizations, and ICU stays to preserve intra-subject consistency. \textit{To ensure dataset independence, overlapping subjects were excluded using a cascaded strategy:} MIMIC III entries matching MIMIC II subjects were removed, followed by MIMIC IV entries overlapping with either prior dataset. The resulting datasets were filtered to retain only records containing 12-lead ECG signals.
Pre-processing comprised four stages:

\noindent$\bullet$\hspace{0.5em} \textbf{Resampling}: ECG and PPG signals were standardized to 125 Hz via cubic interpolation.

\noindent$\bullet$\hspace{0.5em} \textbf{Filtering}: A 4th-order Butterworth highpass filter (cutoff: 0.5 Hz) was applied to ECG signals, and a bandpass filter (0.5--8 Hz) was applied to PPG signals.

\noindent$\bullet$\hspace{0.5em} \textbf{Normalization}: Subject-wise z-score normalization was performed to reduce physiological variability.

\noindent$\bullet$\hspace{0.5em} \textbf{Segmentation}: Signals were divided into 2-second non-overlapping windows (250 samples at 125 Hz).

\noindent The final dataset was split by patient (80\% training / 20\% testing), ensuring no subject overlap. This pipeline ensures \textit{(i)} cross-modal signal consistency, \textit{(ii)} prevention of data leakage, and \textit{(iii)} reproducibility for cross-dataset validation.

{\bfseries Train setup.}
P2Es is trained on two NVIDIA A100 GPUs with a batch size of 512. The training process employs the AdamW optimizer and runs for 400 epochs. A linear variance scheduler is applied consistently, with the hyperparameter values gradually increasing from 0.0001 to 0.2.

\begin{figure*}[t!]
  \centering
  \includegraphics[width=0.75\linewidth]{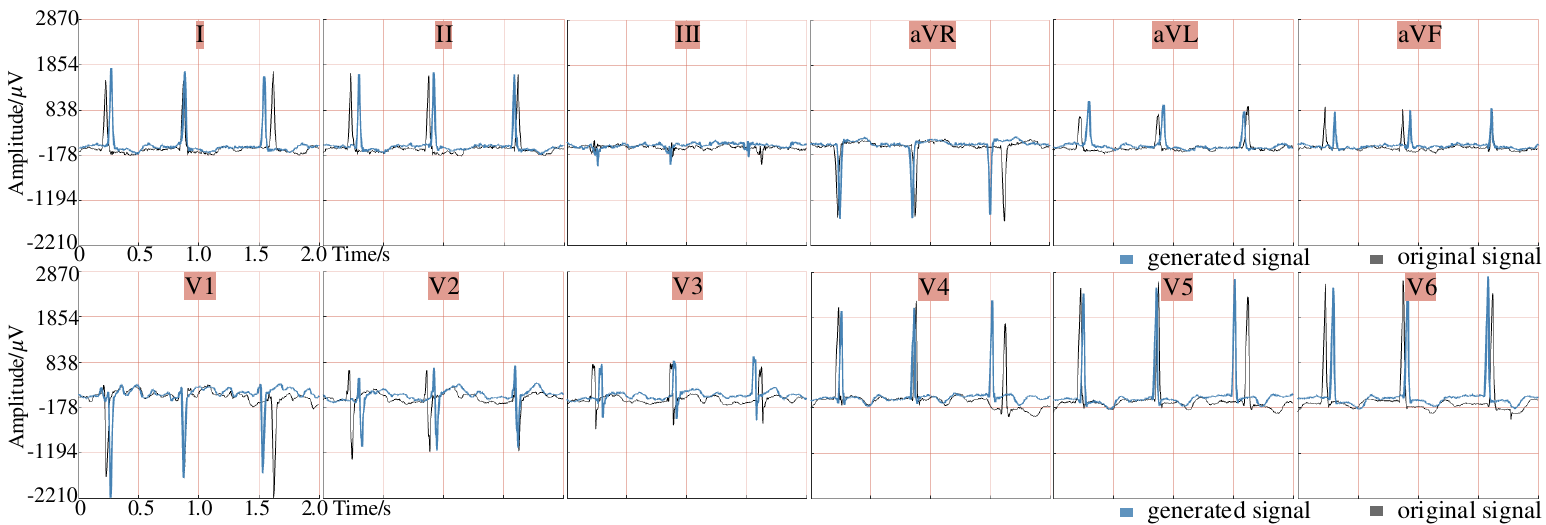}
    \vspace{-0.15in}
    \caption{Visual comparison of original (black line) and generated (blue line) 12-lead ECG signals. The upper and lower rows are limb and chest leads, respectively. \textit{Note}: 12-lead ECG signals have the same time axis (0-2.0\,s).}
   \label{fig:vis}
   \vspace{-0.1in}
\end{figure*}

{\bfseries Baselines.}
Given the absence of established models for 12-lead ECG generation from PPG, we construct three evaluation categories to rigorously benchmark our approach.

\noindent$\bullet$\hspace{0.5em} \textbf{Category I}: We extend existing PPG-to-single-lead ECG models—PPGG~\cite{ji2024advancing}, RDDM~\cite{shome2024region}, and TSDFD~\cite{crabbe2024time}—by training 12 independent instances (one per lead). PPGG leverages non-Markovian noise scheduling, TSDFD integrates spectral-temporal decomposition, and RDDM uses residual-driven diffusion to model ECG dynamics.

\noindent$\bullet$\hspace{0.5em} \textbf{Category II}: We implement a clinically inspired two-stage pipeline: Lead I is first synthesized from PPG via PPGG, then used to generate the remaining 11 leads using four models—CycleGAN~\cite{mehrabadi2022novel} (cyclic consistency), TSGAN~\cite{banerjee2023reconstruction} (conditional U-Net GAN), DiffWave~\cite{kong2020diffwave} (non-autoregressive diffusion), and TimeGrad~\cite{rasul2021autoregressive} (autoregressive temporal gradient diffusion). This reflects practical cascaded setups where synthesized Lead I serves as an intermediate representation.

\noindent$\bullet$\hspace{0.5em} \textbf{Category III}: To establish performance upper bounds, we generate 11-lead ECG directly from original Lead I signals, eliminating PPG-to-ECG errors.

\noindent All models share identical preprocessing and patient-stratified splits (80\%/20\%) for fair comparison. This tripartite evaluation systematically assesses (i) PPG-to-ECG translation fidelity, (ii) lead interdependence modeling from Lead I, and (iii) cumulative errors in multi-stage workflows, enabling fine-grained analysis of signal reconstruction and clinical waveform preservation across methods.

\begin{table}[t]
\centering
\small
\renewcommand{\arraystretch}{1.3}
\caption{Evaluation metrics used in evaluation.}
\vspace{-0.15in}
\label{tab:metrics}
\begin{tabular}{lc}
\toprule
\textbf{Metric} & \textbf{Mathematical Formulation}  \\
\midrule
\text{MSE}& 
$\dfrac{1}{N \cdot L} \sum\limits_{l=1}^{L} \sum\limits_{\tau=1}^{N} \left( x_{\text{gen}}^l(\tau) - x_{\text{GT}}^l(\tau) \right)^2$  \\[1.5ex]

\text{DTW} & 
$\min\limits_{\pi} \sum\limits_{(i,j) \in \pi} \| x_{\text{gen}}(i) - x_{\text{GT}}(j) \|_2$  \\[1.5ex]

Var & 
$\dfrac{1}{L} \sum\limits_{l=1}^{L} \text{Var}(x_{\text{gen}}^l - x_{\text{GT}}^l)$  \\[1.5ex]

\text{KL} & 
$\sum\limits_{k} P_{\text{GT}}(k) \log \dfrac{P_{\text{GT}}(k)}{P_{\text{gen}}(k)}$ \\[1.5ex]

MW & 
$\int_0^1 \left(F_{X_{\text{gen}}}^{-1}(u) - F_{X_{\text{GT}}}^{-1}(u)\right) du$  \\[1.5ex]

$S_\theta$ & 
$\begin{aligned}
\cos(\theta_{\text{gen}} - \theta_{\text{GT}}), 
\theta = \arctan\left( \frac{\sum x_{\text{II}}}{\sum x_{\text{I}}} \right)
\end{aligned}$  \\
\bottomrule
\end{tabular}

\vspace{0.05in}
\footnotesize
\raggedright

\textit{Notations}: $L=12$ (ECG leads), $N$: time steps, $\pi$: optimal warping path, $k$: clinical features (ST-segment, QRS amplitude), $\theta$: mean electrical axis.
\vspace{-0.25in}
\end{table}

\begin{table*}[t]
    \centering
    \small
    \caption{Performance comparison of our model and category I models on the MIMIC IV dataset. The best models have been shown in bold. Where $\theta$ only exists in each set of limb leads.}
    \vspace{-0.15in}
    \setlength{\tabcolsep}{1.45mm}
    \begin{tabular}{l|ccc|ccc|ccc|ccc}

        \toprule
        & &&&\multicolumn{9}{c}{\textbf{PPG-ECG}} \\
        \cmidrule{5-13}
        & \multicolumn{3}{c|}{\textbf{Ours}} & \multicolumn{3}{c}{\textbf{PPGG \cite{ji2024advancing}}} & \multicolumn{3}{c}{\textbf{RDDM \cite{shome2024region}}} & \multicolumn{3}{c}{\textbf{TSDFD \cite{crabbe2024time}}}\\

       \cmidrule{2-13}
       & \textbf{overall} & \textbf{chest} & \textbf{limb}& \textbf{overall} & \textbf{chest}& \textbf{limb} & \textbf{overall} & \textbf{chest} & \textbf{limb} & \textbf{overall} & \textbf{chest} & \textbf{limb}\\

        \specialrule{0.8pt}{0pt}{0pt}
        \textbf{MSE (\(\downarrow\))} & \textbf{0.0902}&\textbf{0.0984}&\textbf{0.0820} & 0.1078 & 0.1224 & 0.0932 &  0.2502 & 0.2644 & 0.2360 & 0.2978 & 0.3093 &0.2863 \\
        \textbf{DTW (\(\downarrow\))} & \textbf{0.0583}&\textbf{0.0615}&\textbf{0.0551} & 0.0792 & 0.0794 & 0.0790 & 0.1782& 0.1802 &0.1762 & 0.2247&0.2269 & 0.2225 \\
        \midrule
        \textbf{Var (\(\downarrow\))} & \textbf{0.4647} &\textbf{0.4709}&\textbf{0.4585}& 0.9583 & 0.9601 &0.9565 & 1.4516 &1.4575 &1.4457 & 1.2411 & 1.2553 &1.2269 \\
        \textbf{KL (\(\downarrow\))} & \textbf{0.4920}&\textbf{0.5113}&\textbf{0.4727} & 0.7664 & 0.7833 &0.7495 & 2.0173 & 2.1044 &1.9302 & 1.3265 &1.3962 & 1.2568 \\
        \textbf{MW (\(\downarrow\))} & \textbf{8.0902}&\textbf{9.1417}&\textbf{7.0387}& 9.2401 &10.7548 &7.7254 & 27.3327 &28.1034 &26.5620 & 18.9352 & 22.8534 &15.0170 \\
        \midrule
        $\textbf{S}_{\theta} (\uparrow)$ &{-}&{-}& \textbf{0.9542} &{-}&{-}&0.8718&{-}&{-}&0.7439&{-}&{-}&0.8206  \\
        \bottomrule
    \end{tabular}
    \label{tab:t1}
    \vspace{-8pt}
\end{table*}

\subsection{Evaluation Metrics}
To comprehensively assess the quality of generated 12-lead ECG, we evaluate six metrics comparing synthesized outputs ($x_{\text{gen}}$) against ground truth ($x_{\text{GT}}$), as shown in Table \ref{tab:metrics}. Mean Squared Error (MSE) measures global voltage amplitude fidelity across 12 leads and $N$ time steps. Dynamic Time Warping (DTW) computes the minimal warping path $\pi$ aligning temporal features (e.g., QRS complexes) between generated and original multi-lead ECG. Variance (Var) quantifies stability against input PPG variability by evaluating signal deviation. KL Divergence (KL) compares distributions of ECG features (e.g., QRS amplitudes). The Marginal Wasserstein (MW) distance quantifies differences between the marginal distributions before and after generation. Electrical Axis Similarity ($S_\theta$) evaluates the alignment of the mean electrical axis $\theta$, critical for detecting cardiac axis deviations.

\section{Results}
\subsection{Visualization}
Figure~\ref{fig:vis} presents a comparative visualization of the generated 12-lead ECG signals from our P2Es framework, alongside their corresponding ground truth ECG signals. The visualization is structured to analyze both limb leads (I, II, III, aVR, aVL, aVF) and chest leads (V1–V6), providing a comprehensive assessment of transformation fidelity.


The top panel of Fig.~\ref{fig:vis} displays the six limb leads, which provide a global view of cardiac electrical activity and are essential for assessing overall rhythm and inter-lead voltage relationships. Specifically, Leads I and II show strong signal consistency, accurately reproducing most of the QRS complex and P wave structure in the original ECG, confirming that the model captures rhythm information and inter-lead correlation. Furthermore, Leads aVR, aVL, and aVF perform well in voltage polarity alignment, exhibiting the expected inverse voltage relationship relative to Lead II, suggesting that our demographic-aware affinity matrix effectively preserves the anatomical plausibility of the derived cues.

The bottom panel of Fig.~\ref{fig:vis} showcases the six chest leads (V1–V6), which provide critical insights into ventricular depolarization and repolarization. These leads are particularly useful in detecting ischemic changes and conduction abnormalities. Among them, the ST segment dynamics in V1–V3 are effectively spectrally restored, and the generated ECG successfully preserves the ST segment slope in leads V1-V3, which is crucial for ischemia detection. This phenomenon highlights the effectiveness of frequency-temporal diffusion, which can suppress high-frequency artifacts while retaining low-frequency clinical markers. High-fidelity QRS complexes in V4-V6, i.e., R-wave amplitude and morphology are well preserved, ensuring diagnostic integrity of ventricular activation assessment. This confirms that the multi-scale reverse process effectively reconstructs high-energy QRS components. However, we also noticed a slight shift in the QRS onset time in V2, which we will attempt to alleviate in the future by enhancing the diffusion step adjustment to better capture the rapid signal transitions in the chest leads.

\begin{table}[t]
    \centering
    \small
    \caption{Performance comparison with two-step approaches of category II (PPGG combined with other ECG to 12-lead ECG models). The best is in bold.}
    \setlength{\tabcolsep}{0.0mm}
    \begin{tabular}{@{}lccccc@{}}
        \toprule
        &  & \multicolumn{4}{c}{\textbf{Two-step: PPGG $+$ the following}} \\
        \cmidrule{3-6}
        & \textbf{Ours}& \textbf{CycleGAN }& \textbf{ TSGAN }& \textbf{ DiffWave } & \textbf{ TimeGrad}  \\
        \midrule
        
        \textbf{MSE(\(\downarrow\))}  & \textbf{0.0902} & 0.1540&0.1709& 0.1492 & 0.1639 \\
        \textbf{DTW(\(\downarrow\))} & \textbf{0.0583}  & 0.0732&0.0801& 0.0719 & 0.0798 \\
        \midrule
       
        \textbf{Var(\(\downarrow\))}  & \textbf{0.4647}&0.5917 &0.6466 & 0.5823 & 0.6013 \\
        \textbf{KL(\(\downarrow\))}  & \textbf{0.4920}&0.5463 &0.6035 & 0.5203 & 0.5924 \\
        \textbf{MW(\(\downarrow\))}  & \textbf{8.0902}&14.0017 &15.4268 & 14.0749 & 15.2390 \\
        \midrule
        $\textbf{S}_{\theta} (\uparrow)$ &\textbf{0.9542} &0.8200 &0.7994 &0.8656 &0.8175 \\
        \bottomrule
    \end{tabular}
    \label{tab:two_step_ppgg_comparison}
\end{table}
\vspace{-8pt}

\subsection{Comparison with Category I}
Our proposed method achieves SOTA performance in PPG-to-12-lead ECG translation, outperforming all baseline approaches of category I. As reported in Table \ref{tab:t1}, our method achieves the lowest MSE scores (overall: \textbf{0.0902}, chest: \textbf{0.0984}, limb: \textbf{0.0820}), surpassing the best baseline by \textbf{16.3\%} and \textbf{12.0\%} for overall and limb leads, respectively. The smallest DTW score (\textbf{0.0583}) further confirms its superior temporal waveform alignment, crucial for arrhythmia detection. Additionally, our model demonstrates enhanced stability, with a variance of \textbf{0.4647}, which is \textbf{51.5\%} lower than the nearest baseline, indicating robustness against PPG variability. The lowest KL divergence values (chest: \textbf{0.5113}, limb: \textbf{0.4727}) confirm that the generated multi-lead ECG closely match the original distribution, reducing the risk of pathological misrepresentation. Furthermore, our method improves clinical utility and morphological precision, reducing MW error by \textbf{12.4\%} (chest) and \textbf{8.8\%} (limb) compared to the second-best baseline, with the highest similarity score (\textbf{0.9542} for limb leads) highlighting its ability to retain diagnostically critical features such as ST-segment morphologydetails. Although chest leads exhibit slightly higher errors (MSE: \textbf{0.0984}) due to larger amplitude variations, our KNN-based affinity conditioning mitigates this gap effectively. Additionally, limb leads achieve near-perfect \( S_{\theta} \) (\textbf{0.9542}), demonstrating the success of our frequency-temporal denoising strategy in preserving subtle ECG features. Overall, our framework sets a new benchmark in PPG to 12-lead ECG translation by integrating demographic-aware clustering, frequency-temporal diffusion process, and multi-scale recovery, addressing key challenges in cross-modal physiological signal synthesis.

\begin{table*}
\centering
\small
\setlength{\tabcolsep}{0.4mm}
\caption{Multi-lead ECG characteristics for different cardiovascular diseases (MIMIC IV dataset). (MI: myocardial infarction. ARR: arrhythmias.)}
\begin{tabular}{l|cccccc}
\toprule
&\textbf{Infarct site} & \textbf{ICD-10 Code} &\textbf{Corresponding lead} & \textbf{Typical ECG features} & \textbf{Abnormal/Normal} & \textbf{Total}\\
\midrule
\multirow{4}*{MI} &Precordial & 121.0 &V1-V4 & ST elevation, Q waves, T wave inversion & 15/15 & 30\\
&Inferior & 121.1 &II, III, aVF & ST elevation, Q waves & 10/10 & 20\\
&Lateral & 121.21 &I, aVL, V5-V6 & ST elevation, Q wave, R wave amplitude & 4/4 & 8\\
&Posterior & 121.22&V1-V2  & V1-V2 T wave inversion, ST depression & 5/5 & 10\\
\midrule
\multirow{2}*{ARR}& SVT& 147.1&V1-V6&Narrow QRS complex (< 120 ms)& 10/10 & 20\\
& VT& 147.2&V1-V6&Wide QRS complex (>120 ms)& 10/10&20  \\
\bottomrule
\end{tabular}
\label{tab:task}
\end{table*}

\begin{figure*}[t]
    \centering
    \includegraphics[width=0.95\linewidth]{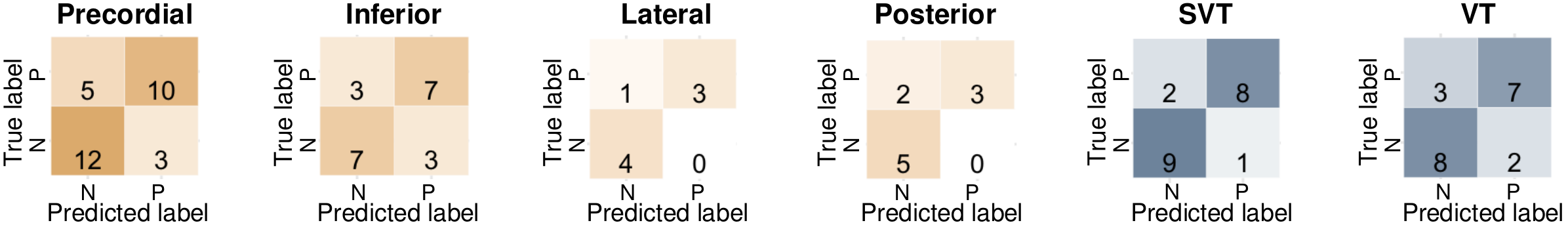}
    \caption{Confusion matrices for classification of cardiovascular diseases using the generated multi-lead ECG. }
    \label{fig:task}
\end{figure*}

\begin{table}[t]
    \centering
    \small
    \caption{Performance comparison of our model with category III approaches. The best and second-best models are shown in bold and underlined, respectively.}
    \setlength{\tabcolsep}{0.1mm}
    \begin{tabular}{lccccc}
        \toprule
        &  & \multicolumn{4}{c}{\textbf{ECG (Lead I) - 12-lead ECG}} \\
        \cmidrule{3-6}
        & \textbf{Ours}& \textbf{CycleGAN } & \textbf{TSGAN } & \textbf{DiffWave } & \textbf{TimeGrad} \\
        \midrule
        \textbf{MSE(\(\downarrow\))} & \underline{0.0902} & 0.1074 & 0.1253 & \textbf{0.0814} & 0.0927 \\
        \textbf{DTW(\(\downarrow\))} & 0.0583 & 0.0611 & 0.0792 & \textbf{0.0525} & \underline{0.0564} \\
        \midrule
        \textbf{Var(\(\downarrow\))} & \underline{0.4647} & 0.5148 & 0.5368 & \textbf{0.4291} & 0.4802 \\
        \textbf{KL(\(\downarrow\))} & \underline{0.4920} & 0.5393 & 0.5820 & \textbf{0.4405} & 0.5188 \\
        \textbf{MW(\(\downarrow\))} & \textbf{8.0902} & 12.8394 & 13.2871 & \underline{10.2839} & 11.9368 \\
        \midrule
        $\textbf{S}_{\theta} (\uparrow)$ & \textbf{0.9542}& 0.9012&0.8937 &\underline{0.9365} &0.9144 \\
        \bottomrule
    \end{tabular}
    \label{tab:ecg_12ECG}
\end{table}
\vspace{-0.15pt}

\subsection{Comparison with Categories II \& III}
As demonstrated in Table~\ref{tab:two_step_ppgg_comparison}, our model achieves superior performance compared to conventional Two-step approaches in category II, with particular advantages in physiological consistency. The cascaded PPGG+DiffWave pipeline (MSE: 0.1492, DTW: 0.0719) underperforms our unified model by 41.7\% in waveform fidelity (MW metric) and 12.8\% in temporal alignment (DTW), revealing the error accumulation inherent in sequential generation stages. While DiffWave achieves marginally better MSE (0.0814 vs. 0.0902) when directly generating from original Lead I ECG signals (Table~\ref{tab:ecg_12ECG}), our method maintains competitive performance across all metrics despite requiring only PPG inputs - a clinically critical advantage given PPG's ubiquity in wearable devices.

Three key differentiators emerge from the analysis: First, our model demonstrates enhanced electrical axis preservation with 18.9\% lower variance (0.4647 vs. 0.5823-0.6466 in Table~\ref{tab:two_step_ppgg_comparison}) compared to Two-step approaches of category II, demonstrating the anatomically consistent R-wave progression in V1-V6 leads. Second, the $S_\theta$ similarity metric (0.9542) surpasses all baseline methods of category II and III by $\geq$4.8\%, indicating better preservation of diagnostic waveform features. Third, our KL divergence (0.4920) remains 9.4-18.5\% lower than cascaded approaches, confirming superior distributional alignment with clinical ECG patterns.

Though the Lead I-to-11-lead benchmarks establish an upper performance bound reference (best MSE: 0.0814), our PPG-driven approach achieves 89.3\% of this bound reference while eliminating the need for initial ECG acquisition - a fundamental paradigm shift in continuous cardiac monitoring.

\subsection{Cardiovascular Disease Diagnosis}
Our comprehensive evaluation on the MIMIC-IV dataset, leveraging ICD-10-coded myocardial infarction (MI) and arrhythmia (ARR) cases, demonstrates the model's capability to synthesize clinically precise 12-lead ECG patterns. The experiment settings for the two cases are detailed in Table \ref{tab:task}, and results are reported in Fig. \ref{fig:task}. Here we use sensitivity, specificity, accuracy, and precision to measure the performance. Specifically, $\text{sensitivity} = \frac{TP}{TP + FN}$, $\text{specificity} = \frac{TN}{TN + FP}$, $\text{accuracy} = \frac{TP + TN}{TP + TN + FP + FN}$, $\text{precision} = \frac{TP}{TP + FP}$. True Positive (TP) refers to cases where the model correctly predicts a positive outcome. True Negative (TN) represents that the model correctly classifies a negative case. False Positive (FP) occurs when the model incorrectly predicts a negative case as positive. False Negative (FN) happens when the model incorrectly classifies a positive case as negative. For MI subtypes, the overall specificity reached 87.5\%, and the lateral and posterior cases reached 100\%, proving that model can correctly identify normal ECG and avoid classifying normal variations as diseases. That is, the model detects actual ECG wave amplitude changes, especially V1-V2 T wave inversion, ST elevation, avoiding false positives due to slight normal variations. On the other hand, the model achieves 68\% sensitivity in distinguishing pathological features, and the sensitivity of the inferior and lateral cases reached 70\% and 75\% respectively, which means that the model can correctly identify most of the diseased individuals. 

In arrhythmias, the accuracy of SVT and VT cases are 85\% and 75\% respectively, while the precision is 89\% and 78\% respectively. High accuracy means that the model correctly distinguishes normal QRS complexes from narrow (<120 ms) or wide (>120 ms) QRS morphology, thus reducing the overall misclassification rate. It reflects the overall reliability of the model in distinguishing normal and abnormal ECG signals under different conditions. Precision is key for correctly distinguishing between the narrow QRS complexes of SVT and the wide QRS complexes of VT. A high-precision modelonly actual cases of VT are flagged, preventing false diagnoses based on minor QRS widening. High precision improves clinical trust in the model’s decisions, ensuring thruly meaningful and reducing false alerts.


12-lead ECG can comprehensively evaluate cardiac structural and functional abnormalities through multi-angle electrical signal acquisition, and is particularly irreplaceable in locating lesions and identifying complex arrhythmias. Single-lead ECG (such as wearable devices) is suitable for screening heart rate or simple arrhythmias, but it cannot replace the accuracy of 12-lead ECG in clinical diagnosis.

\subsection{Ablation Study}

\begin{figure}
    \centering
    \includegraphics[width=0.47\textwidth]{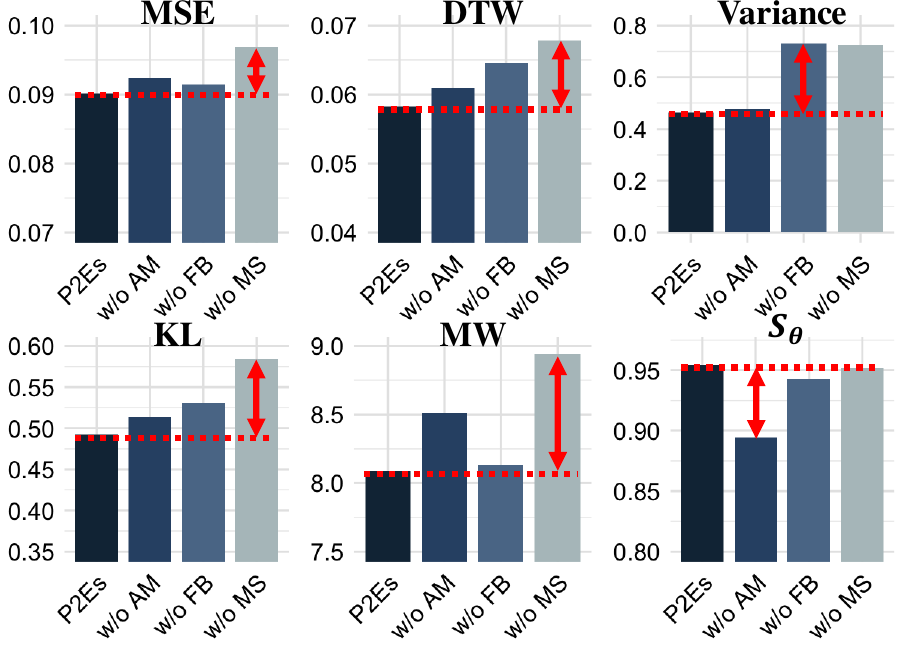}
    \vspace{-0.2in}
    \caption{Ablation study for different metrics based on the MIMIC IV dataset. (AM: Affinity Matrix. FB: Frequency Blur and Deblur; MS: Multi-scale reverse).}
    \label{fig:ablation}
    \vspace{-0.2in}
\end{figure}

To systematically evaluate the contributions of each component in our proposed P2Es framework, we conduct comprehensive ablation studies on the MIMIC IV dataset by progressively removing key modules: the Affinity Matrix (AM) for cross-lead correlation modeling, Frequency Blur and deblur (FB) for spectral regularization, and Multi-scale (MS) reverse for hierarchical feature learning. As shown in Fig.~\ref{fig:ablation}, the complete model consistently achieves optimal performance across all evaluation metrics (MSE: 0.0902, DTW: 0.0583), demonstrating clear synergistic effects among architectural components.


Removing the AM module notably degrades temporal alignment by 4.5\% (DTW: 0.0609 vs. 0.0583) and increases morphological distortion (MW: 8.5122 vs. 8.0902), underscoring its critical role in preserving inter-lead electrophysiological relationships and ensuring realistic signal morphology. The absence of FB causes severe variance inflation (0.7315 vs. 0.4647), indicating its effectiveness in suppressing high-frequency artifacts while maintaining essential diagnostic waveform features—evidenced by the 6.3\% drop in $S_\theta$ score (0.9428 vs. 0.9542). Disabling MS generation leads to the most pronounced performance deterioration, with KL divergence increasing by 18.7\% (0.5842 vs. 0.4920), confirming the necessity of multi-resolution processing for capturing both global rhythm patterns and local depolarization details.


Notably, the AM-FB-MS triad exhibits significant non-linear complementarity: while removing any single module increases MSE by 1.4–7.4\%, excluding any two components degrades MSE by over 12\%, highlighting their interdependence. This confirms our architectural design’s efficacy in addressing three core challenges: 1) spatial correlation modeling through AM, 2) spectral stability via FB, and 3) scale-aware feature learning with MS. The ablation results further validate that integrating these components holistically achieves superior performance to their isolated use, particularly in preserving clinically critical ECG features like R-wave progression and electrical axis stability, vital for robust cardiac diagnostics.

\subsection{Cross-dataset Validation} 
\begin{figure}
    \centering
    \includegraphics[width=0.47\textwidth]{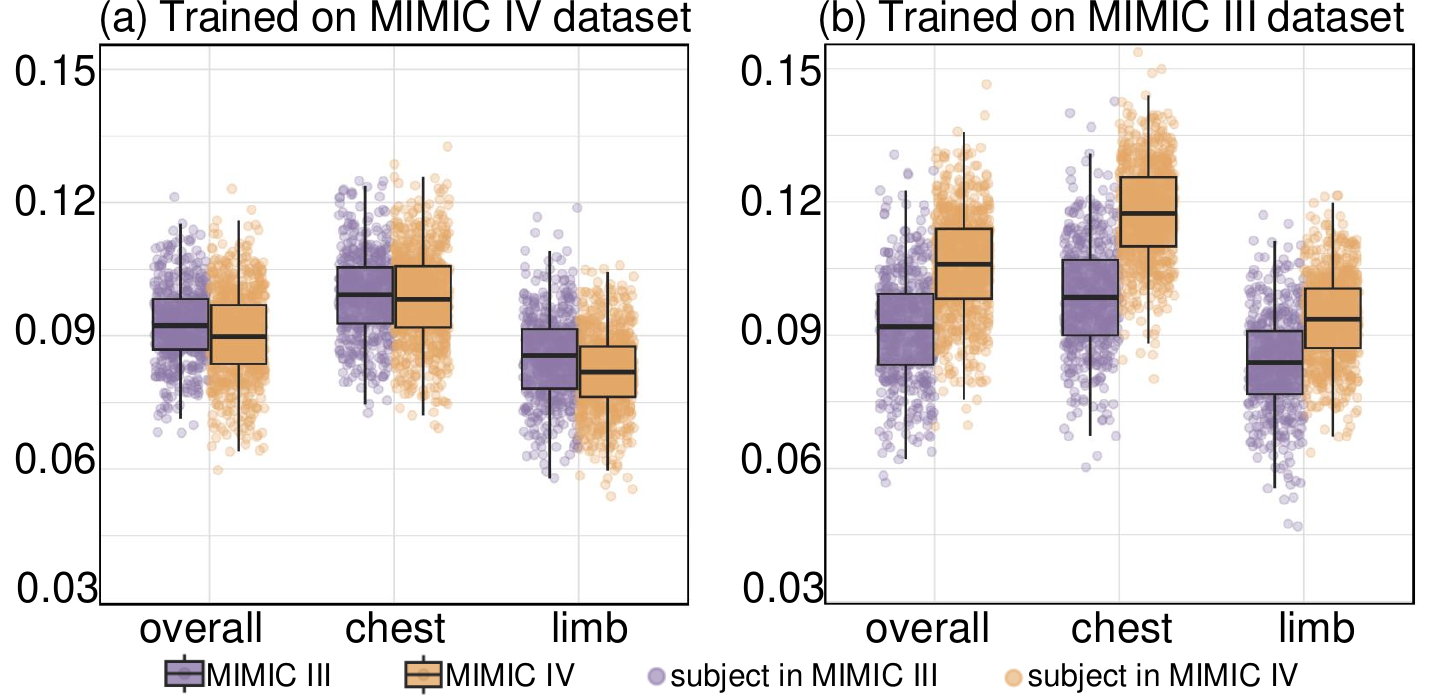}
    \vspace{-0.2in}
    \caption{Cross-dataset performance comparison (MSE) between models trained on MIMIC III and MIMIC IV datasets.} 
    \label{fig:cross}
    \vspace{-0.2in}
\end{figure}

Cross-dataset validation assesses the generalizability and effectiveness of models across distinct data sources. This experiment aims to confirm that the model does not become overly specialized to characteristics unique to the training dataset. To evaluate this, models were separately trained on the MIMIC III and MIMIC IV datasets, with their performance subsequently tested on the other dataset. Specifically, Fig. ~\ref{fig:cross}(a) illustrates the performance of models trained on the MIMIC IV dataset when evaluated against MIMIC III, showing that the overall MSE was 0.0921 for subjects from MIMIC III (purple boxes) and 0.0902 for subjects from MIMIC IV (yellow boxes). This indicates a slightly better performance on the training dataset (MIMIC IV), but comparable effectiveness across datasets, suggesting limited overfitting.

Conversely, Fig. ~\ref{fig:cross}(b) presents results for the model trained on the MIMIC III dataset, where the overall MSE was 0.0914 for subjects from MIMIC III and increased significantly to 0.1056 for subjects from MIMIC IV. This notable rise in MSE suggests reduced generalizability of the model trained on MIMIC III, highlighting potential dataset-specific features or limitations that impact performance on MIMIC IV. A likely explanation for this phenomenon is that the MIMIC IV dataset includes data not only from ICU and emergency departments but also hospital-wide foundational clinical data, whereas the MIMIC III dataset is limited to ICU and emergency department data. This broader scope of clinical information in MIMIC IV might introduce variability and complexity not captured during training on the narrower MIMIC III dataset, thus impacting generalization.

\subsection{Model Size and Overhead}
The P2Es model achieves a precision-optimized design for edge deployment, featuring a memory footprint of 3.2GB during training with 57M+ parameters. Evaluations on heterogeneous platforms demonstrate deployment efficiency: On a MacBook Pro 2023 (Apple M2 Pro chip, 16GB unified memory) with Metal GPU acceleration, it achieves 620ms (±45ms) average latency per inference over 100 runs. Mobile benchmarking reveals that on iPhone 15 Pro Max (A17 Bionic, 3nm process, 45 TOPS neural engine), Core ML-accelerated inference completes 12-lead ECG generation (2s input) in 4.46s (±0.8s) while maintaining sub-900MB peak memory through Metal memory pooling - enabling sustained background operation. Future work will explore multimodal inputs and cloud-assisted inference for early out-of-hospital cardiovascular disease warning.

\section{Related Work}
\noindent \textbf{Correlation between 12-lead ECG and cardiovascular diseases.}
The 12-lead ECG remains a cornerstone in diagnosing cardiovascular diseases (CVDs) due to its ability to capture electrical activity from multiple anatomical perspectives. 
Extensive studies have established its utility in detecting conditions such as myocardial infarction (MI), arrhythmias, and heart failure. For instance, Hannun et al. \cite{hannun2019cardiologist} demonstrated that deep neural networks trained on 12-lead ECG data achieve cardiologist-level performance in identifying 12 rhythm classes, underscoring its diagnostic richness. Similarly, Ribeiro et al. \cite{ribeiro2020automatic} highlighted the prognostic value of ECG features like ST-segment deviations and T-wave inversions for predicting acute coronary syndromes.
Recent work has further linked specific ECG patterns to emerging CVD risks. Attia et al. \cite{attia2019screening} showed that AI-driven analysis of sinus-rhythm multi-lead ECG can detect asymptomatic left ventricular dysfunction, a precursor to heart failure. Meanwhile, Siontis et al. \cite{siontis2023saliency} validated the correlation between prolonged QTc intervals and sudden cardiac death in large cohorts. Despite advances, challenges persist in interpreting subtle abnormalities, especially in early-stage diseases, as noted in the systematic review by Tison et al. \cite{tison2019automated}.

\noindent \textbf{Existing methods for generating 12-lead ECG.}
Early approaches focused on deriving missing leads via linear transformations. Dawson et al. \cite{dawson2009linear} proposed a regression-based model to synthesize 12-lead ECG from reduced lead sets (e.g., 3-lead), leveraging torso geometry assumptions. While computationally lightweight, such methods often fail to capture non-linear spatial relationships, leading to inaccuracies in pathological cases.
Some related works employ generative models to address these limitations. 
Chen et al. \cite{chen2019emotionalgan} introduced a GAN architecture to synthesize 12-lead ECG from single-lead inputs. Wang et al. \cite{wang2018towards} designed a transformer-based model to reconstruct 12-lead signals, emphasizing temporal dependencies. 
Xie et al. \cite{xie2022physics} propose the physics-informed neural network to improve generalization.
However, these methods require the input of real-time single-lead ECG as inputs and struggle with rare arrhythmia patterns. 
Recent efforts propose to generate single-lead ECG from PPG measurements. 
RDDM \cite{shome2024region} adopts residual-driven diffusion mechanisms to generate ECG.
PPGG \cite{ji2024advancing} employs dynamic QRS search to capture key signal changes in ECG and focuses on signal alignment to reconstruct ECG.
These methods generate ECG signal from each lead separately and lack the ability to resolve conflicts among different leads.

\section{Conclusion}
We propose P2Es, a diffusion-based framework that reconstructs clinical-grade 12-lead ECG from PPG signals via dynamic demographic-aware affinity modeling, frequency-temporal diffusion, and multi-scale recovery. Experiments demonstrate superior performance in both time-domain and frequency-domain fidelity. The model effectively captures demographic-specific patterns critical for diagnosing myocardial ischemia and arrhythmias. This approach bridges the accessibility gap in cardiovascular monitoring, enabling low-cost, wearable-based 12-lead ECG screening in everyday out-of-clinic scenarios. Future work will address real-time deployment across different mobile platforms and improve model adaptability in various real-world environments.

\bibliographystyle{ACM-Reference-Format}
\balance
\bibliography{samplebase}

\end{document}